\documentclass[10pt,twocolumn,letterpaper]{article}

\usepackage[pagenumbers]{cvpr} 

\usepackage{microtype}

\usepackage{multirow}

\definecolor{good}{HTML}{1BAF7A}
\definecolor{bad}{HTML}{EB6834}

\definecolor{cvprblue}{rgb}{0.21,0.49,0.74}
\usepackage[breaklinks,colorlinks,allcolors=cvprblue]{hyperref}  
\usepackage{orcidlink}  

\def\paperID{*****} 
\def\confName{CVPR}
\def\confYear{2026}

\title{Wind on Trees: Testing Physical Grounding in Dynamic 4D Gaussian Splatting}

\author{%
Weiying Chen\,\orcidlink{0009-0007-9212-0745}\\
University of Alberta\\
{\tt\small weiying3@ualberta.ca}
\and
Edmond Lou\,\orcidlink{0000-0002-7531-8377}\\
University of Alberta\\
{\tt\small elou@ualberta.ca}
}

\begin{document}
\maketitle

\begin{figure*}[!t]
  \centering
  \includegraphics[width=\linewidth]{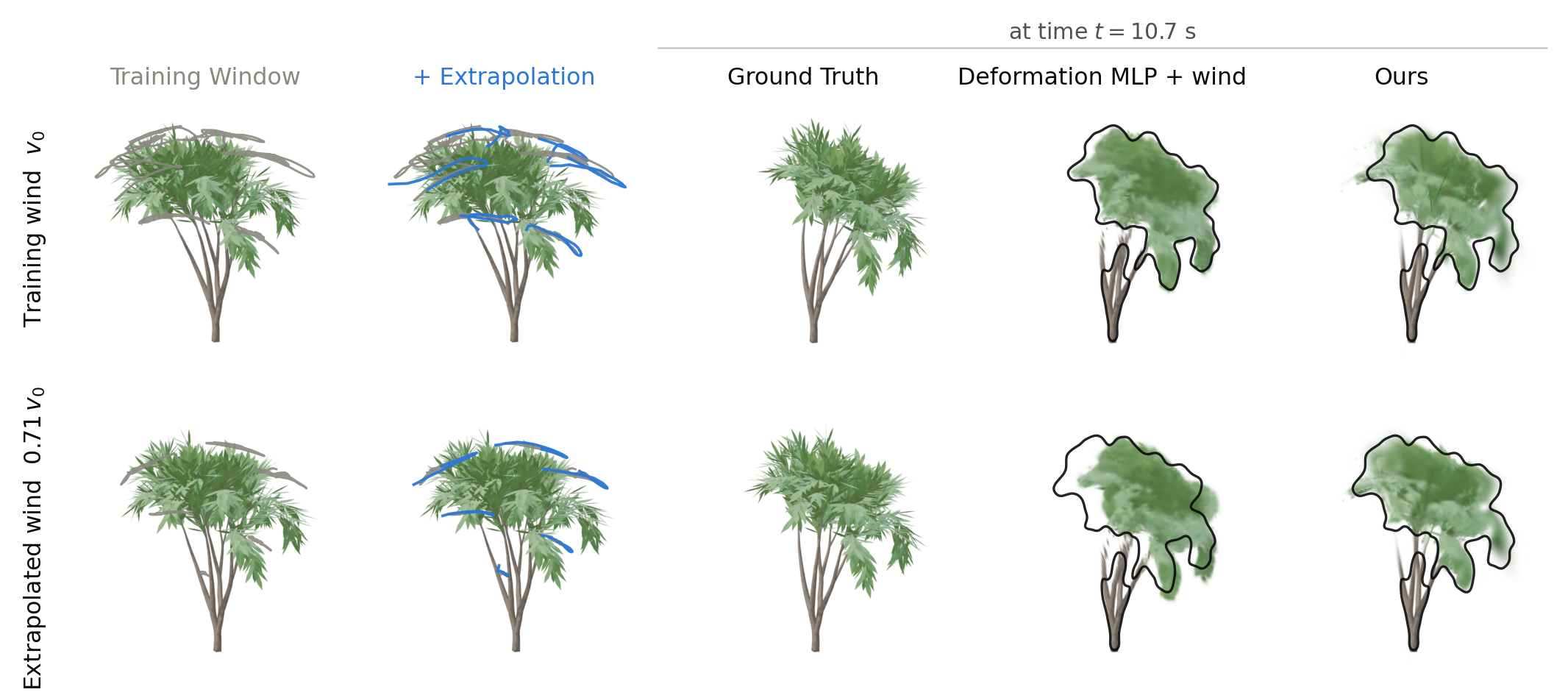}
  \caption{\textbf{Does a reconstruction recover physics, or replay an animation?} Both models are trained on the top row's wind only. \emph{Columns 1--2:} branch-tip trajectories tracked through the ground-truth frames, over the 300-frame training window (grey) and continuing into the 50-frame temporal-extrapolation tail (blue). \emph{Columns 3--5:} a single instant from \emph{inside} the training window, with the ground-truth canopy outline overlaid on both predictions, so only the wind differs between rows. On the training wind (top), both models sit on the outline: 5.4\,px and 2.7\,px of canopy-centroid error, and no photometric comparison separates them. Extrapolate the wind to $0.71v_0$ (bottom) and the Deformation MLP drifts off the outline (32.6\,px) while ours stays on it (0.1\,px). \emph{Deformation MLP\,+\,wind} is a Fourier-encoded MLP deformation field over canonical position and time~\cite{Yang2024Deformable}, additionally conditioned on the preceding 6\,s of observed wind subsampled to 48 values, and trained under the same photometric loss as ours.}
  \label{fig:teaser}
\end{figure*}

\begin{abstract}
Monocular reconstruction of wind-driven vegetation is severely underconstrained: motion along the viewing direction is largely unobservable, a moving canopy offers few reliable correspondences, and nearly the entire scene is dynamic, providing little static reference. Directly-learned deformation fields in 4D Gaussian Splatting therefore optimize photometric consistency rather than recover the motion that produced it. We replace that field with a physically parameterized deformation prior: one damped harmonic oscillator per rigid part, driven by the observed wind and integrated by differentiable RK4, supervised photometrically alone. To test whether such a prior is physically grounded rather than merely well fit, we build a controlled synthetic testbed of three procedurally generated trees spanning an order of magnitude in skeleton complexity, whose per-part natural frequency follows from its own geometry and whose damping ratio is a fixed constant, both held out of training. On it, we measure held-out views, temporal extrapolation, zero-shot transfer to unseen wind speeds, and recovery of the physical parameters themselves. The prior costs appearance fidelity on in-distribution views and extrapolates markedly better outside the training window and the training wind, while parameter recovery is far weaker than it first appears: frequency recovery survives an untrained null control on only the sparsest of the three trees, and damping is not recovered at all.
\end{abstract}

\section{Introduction}
\label{sec:intro}

In an outdoor digital twin, or a game scene that responds to weather, what matters about a tree is often not the gust that was filmed but how it would move under a wind that was never recorded. Reconstructing a wind-driven tree from a single fixed camera is one of the most underconstrained instances of 4D reconstruction: motion along the viewing direction is largely unobservable, a moving canopy offers few reliable correspondences, and almost the entire scene moves at once, leaving little static content to anchor scale or pose. Dynamic Gaussian Splatting represents these scenes with a deformation field over canonical position and time~\cite{Yang2024Deformable,Wu20244DGS}. Photometric error is the field's only constraint, which is enough inside the window it was fit to and says nothing outside it: beyond the last training frame, or under a wind speed never recorded, a time-indexed field has nothing left to do but replay. \cref{fig:teaser} shows the consequence, where two models that no photometric comparison separates on the training wind come apart as soon as the wind changes.

Physically-driven vegetation is well developed in the generative direction. Graphics has animated trees from per-branch oscillators and modal bases for decades~\cite{Chuang2005Stochastic,Habel2009Physically,Diener2009Wind,Stam1997Stochastic}, and recent work synthesizes plausible tree motion with learned models~\cite{Li2026DynamicTree} or couples simulators to Gaussians with material parameters supplied by hand~\cite{Xie2024PhysGaussian,Lin2025OmniPhysGS}. The inverse direction is largely absent for trees: recovering, from video of one moving, the structural parameters that explain the motion. Part of why that gap persists is that there is no released data carrying labelled physical parameters alongside rendered images, against which recovery could be checked. Even the standard procedural source of tree motion offers none: Blender's Sapling add-on, the common implementation of Weber and Penn's model~\cite{WeberPenn1995}, drives each bone with fixed-frequency sinusoids whose frequency comes from a global animation-speed setting rather than from the branch's own natural frequency, with no damping ratio and a constant wind slider in place of a measured signal. In this paper, we propose a physically parameterized deformation prior for wind-driven trees, together with a testbed on which what it recovers can be scored against physical ground truth.

Our contributions are three-fold:
\begin{itemize}
  \setlength{\itemsep}{2pt}
  \item \textbf{A testbed with recoverable physics.} We build three procedural trees spanning 22 to 562 animated bones, whose per-part natural frequency $\omega_p$ follows from geometry and whose damping ratio $\zeta_p$ is a fixed constant, both held out of training, with novel views, a temporal-extrapolation tail, and three wind conditions.
  \item \textbf{A physically parameterized deformation prior.} We replace the learned field with per-part damped oscillators driven by the observed wind and integrated by differentiable RK4, supervised photometrically alone; this trades in-distribution fidelity for extrapolation.
  \item \textbf{An evaluation with null controls.} We observe that part-to-bone matching is degenerate on dense skeletons, so frequency recovery holds only on the sparse tree; damping is not recovered.
\end{itemize}

\section{Related Work}
\label{sec:related}

\paragraph{Dynamic and Deformation-Based Gaussian Splatting.}
Dynamic Gaussian Splatting pairs canonical 3D Gaussians with a deformation that depends on time. Deformable-3DGS learns that deformation as an MLP over encoded canonical position and time, predicting a per-Gaussian offset in position, rotation and scale~\cite{Yang2024Deformable}; 4D-GS encodes space and time together in a multi-resolution plane structure decoded by a small MLP~\cite{Wu20244DGS}. Dynamic 3D Gaussians instead gives the motion structure, optimizing a free position and rotation per Gaussian per timestep from multi-view video, regularized so that neighbours move as near-rigid groups~\cite{Luiten2024Dynamic3DGS}. SC-GS makes the parts explicit: sparse control points carry learned time-varying rigid transforms and skin the dense Gaussians to them~\cite{Huang2024SCGS}. Rigid-part decomposition is thus established and is not what we claim; what differs is where a part's transform comes from. In all of them, it is a free-form function of time, fit to the recorded frames by photometric error alone, so none defines what should happen after the last one. In ours, it is the response of a damped oscillator with per-part physical parameters to an externally observed cause, the measured wind speed.

\paragraph{Physics-Informed Gaussian Splatting and System Identification.}
A simulator does not by itself close the gap between fitting an observed motion and identifying the mechanics behind it. PhysGaussian couples a Material Point Method simulator to 3D Gaussians~\cite{Xie2024PhysGaussian}; its hand-specified material is single and homogeneous, so nothing is recovered from observation. OmniPhysGS models heterogeneous per-Gaussian materials~\cite{Lin2025OmniPhysGS}, but is generative: dynamics are synthesized under text or video-diffusion supervision. Two works do recover material from video, differentiating through a simulator under photometric supervision. Spring-Gaus fits a per-anchor stiffness field to calibrated multi-view video of a dropped elastic object, with mass and damping held fixed~\cite{Zhong2024SpringGaus}. PhysDreamer fits a Young's modulus field through a differentiable MPM simulator so that renderings match a video-diffusion clip, noting that no measured material ground truth exists for its objects~\cite{Zhang2024PhysDreamer}. Both recover an elastic material field rather than per-part oscillator parameters, from unobserved excitation, with nothing independent to score against. DynamicTree generates tree motion as a sparse voxel spectrum from images of a static tree~\cite{Li2026DynamicTree}, observing no motion at inference, taking no wind input, and adopting its modal mass, damping and stiffness from prior work. Its synthetic 4DTree set shares our Weber and Penn base~\cite{WeberPenn1995}, but is described only as animated meshes with vertex motion, so as distributed it offers neither rendered images nor labelled mechanical parameters.

Outside the vegetation setting, PAC-NeRF identifies material parameters from multi-view video by differentiating through a continuum simulator~\cite{Li2023PACNeRF}. It recovers a few global constants for a single object rather than a distribution of per-part parameters over an articulated skeleton, and its target is the simulator's own input rather than a quantity derived from geometry. Closest in spirit is \emph{Seeing the Wind from a Falling Leaf}~\cite{Gao2025SeeingWind}, which recovers an invisible external force from the motion of one falling leaf. We treat the wind as a known and observable input instead, and recover the tree's own intrinsic parameters, natural frequency and damping, across many coupled parts.

\paragraph{Tree Dynamics and Wind.}
Tree biomechanics has long modelled a swaying tree or branch as a forced damped harmonic oscillator~\cite{James2014Dynamics,Kerzenmacher1998Model}. Its natural frequency follows a cantilever-beam scaling in radius and length~\cite{Rodriguez2008Scaling,Jackson2019Architectural,Moore2004Sway}. Measured damping ratios are small: Jackson et al.\ report $8.6 \pm 2.2\%$ for foliated trees in summer~\cite{Jackson2019Architectural}, consistent with earlier crown measurements~\cite{Moore2005Crown}. Graphics has independently converged on the same lumped description for animation: per-branch oscillators driven by a turbulence spectrum~\cite{Chuang2005Stochastic,Habel2009Physically}, or modal bases projected onto wind~\cite{Stam1997Stochastic,Diener2009Wind}. Both literatures use these models \emph{generatively}, to produce plausible motion from assumed parameters. We invert the direction of use: the same equations supply a ground truth against which a reconstruction method's recovered parameters can be scored, which is what makes the testbed falsifiable rather than merely plausible.

\section{Method}
\label{sec:method}

\begin{figure*}[t]
  \centering
  \includegraphics[width=\linewidth]{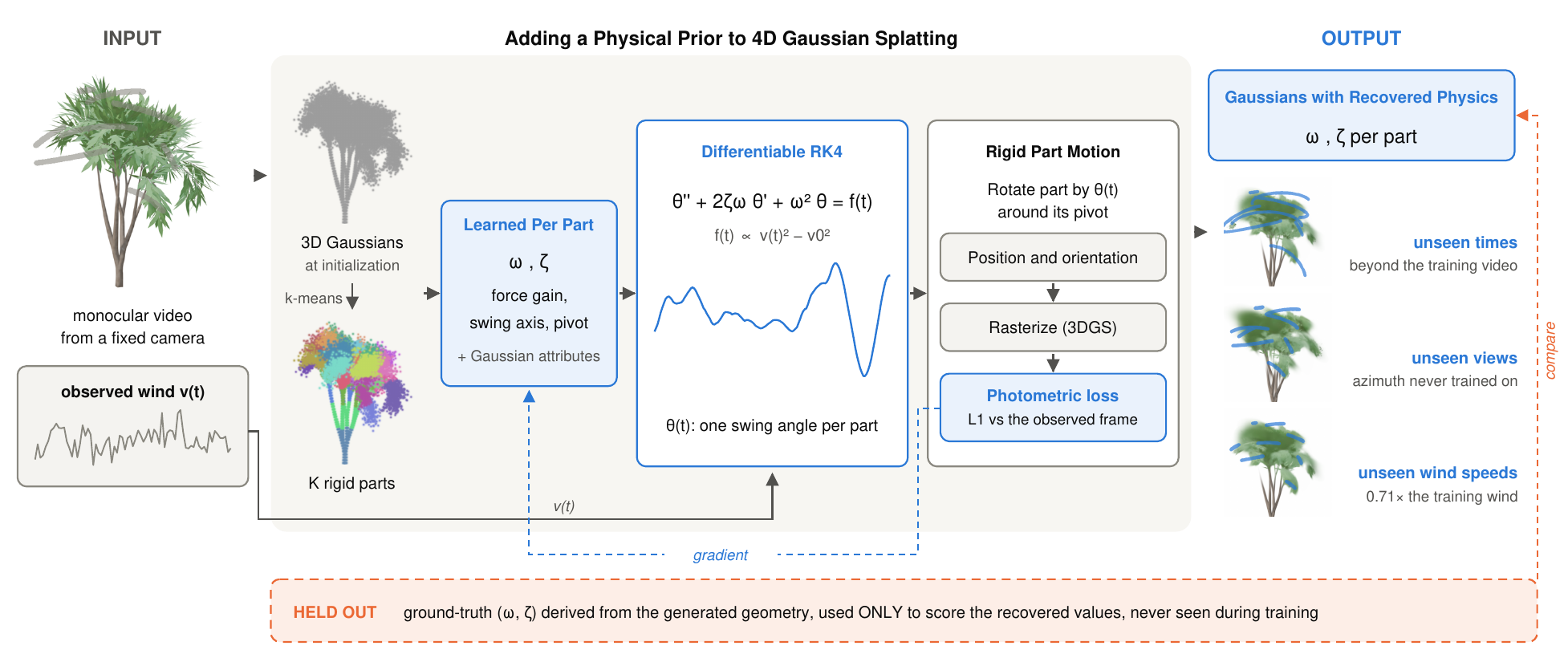}
  \caption{\textbf{Overview of Our Method.} Our method takes two inputs: a monocular video from a fixed camera, and the wind speed $v(t)$ recorded alongside it. Canonical Gaussians are hard-assigned to $K$ rigid parts by $k$-means on position (colours are parts), and each part learns $\omega$, $\zeta$, a force gain, a swing axis, and a pivot. The wind drives one damped oscillator per part, which differentiable RK4 integrates into a swing angle $\theta(t)$. Each part is then rigidly rotated by $\theta(t)$ around its pivot, with the rotation applied to Gaussian position \emph{and} orientation, and rasterized. The wind is a measurement, not supervision: the only training signal is photometric, and nothing supervises $\omega$ or $\zeta$. Trails are swing paths of tracked canopy points, grey for the observed window and blue for prediction at unseen times, views, and wind speeds. Ground-truth $(\omega,\zeta)$ is held out: it only scores the recovered values, and never enters training.}
  \label{fig:pipeline}
\end{figure*}

\subsection{Synthetic Testbed Construction}
\label{sec:method:testbed}

\paragraph{Geometry.} We generate armature-based procedural trees using Weber and Penn's model~\cite{WeberPenn1995}, with the mesh bound to bones via vertex groups one-to-one, so every mesh vertex traces back to the bone whose rigid motion explains its displacement. The Gaussians themselves inherit nothing but initial positions from that mesh (\cref{sec:method:prior}): the vertex-to-bone correspondence is discarded, so no per-Gaussian bone label exists and learned parts must instead be matched to bones by their dynamics (\cref{sec:results:omega}). We instantiate three trees spanning distinct morphologies, named throughout by skeleton complexity: \emph{sparse}, a two-branch fork (22 animated bones); \emph{medium}, a round canopy (224); and \emph{dense}, a larger, bushier canopy (562). Skeleton complexity thus varies by more than an order of magnitude across the testbed.

\paragraph{Ground-Truth Physics.} Rather than hand-assigning oscillation parameters, each bone's natural frequency follows from its own generated geometry via a cantilever first-mode scaling law,
\begin{equation}
    \omega_p \;\propto\; \frac{r_p}{L_p^2}\sqrt{E/\rho},
    \label{eq:cantilever}
\end{equation}
with $r_p,L_p$ the bone's radius/length and $E/\rho$ a fixed material constant, auto-calibrated per tree. \cref{eq:cantilever} is the first-mode scaling for a clamped-free Euler beam of solid circular section~\cite{Blevins1979Formulas}, applied at branch-segment level following Rodriguez et al.~\cite{Rodriguez2008Scaling}; the same $r/L^2$ form has been validated against field measurements and finite-element models of laser-scanned trees at whole-tree level~\cite{Jackson2019Architectural,Moore2004Sway}. Thin, distal bones therefore oscillate faster than thick, proximal ones without manual per-bone assignment. We note the approximation is reported to hold better for conifers than for broadleaves~\cite{Jackson2019Architectural}.

\paragraph{Forcing.} Each bone's angle follows a damped harmonic oscillator,
\begin{equation}
    \ddot\theta_p + 2\zeta_p\omega_p\dot\theta_p + \omega_p^2\theta_p = f_p(t),
    \label{eq:oscillator}
\end{equation}
the standard lumped form used for wind-driven tree motion~\cite{James2014Dynamics,Chuang2005Stochastic,Habel2009Physically}, with linear viscous damping as is conventional in this literature. Forcing follows a quadratic drag law, $f_p(t) \propto v(t)^2 - v_0^2$~\cite{deLangre2008Wind,Diener2009Wind}. Damping is held at $\zeta_p = 0.08$ for every bone.

Excitation is broadband, as atmospheric turbulence is~\cite{Kaimal1972Spectral}, so that every bone across the $\sim$4.3$\times$ span of natural frequencies (0.35--1.5\,Hz) is driven near its own resonance; a narrow spectrum leaves high-frequency bones undriven and their $\omega_p$ structurally unrecoverable by any method. Our flat 0.1--8\,Hz band is chosen for parameter identifiability, not for meteorological realism: a real surface-layer spectrum decays as $f^{-5/3}$, and our upper limit follows what the cantilever law predicts for thin, high-order branches rather than measured whole-tree frequencies, which are well below 1\,Hz. A per-bone force-gain compensation keeps response amplitude comparable across the band, counteracting the quasi-static $1/\omega^2$ scaling.

\paragraph{Scope of the Damping Constant.} That value sits inside the $8.6 \pm 2.2\%$ measured for foliated trees in summer~\cite{Jackson2019Architectural} and at the lower bound of the range for unpruned Douglas-fir~\cite{Moore2005Crown}, both whole-tree first-mode measurements, so applying it per bone extrapolates in scale. It is also an \emph{effective} structural-plus-aerodynamic ratio, not the material damping of wood: writing the drag on $v^2-v_0^2$ rather than on the relative velocity $\tfrac{1}{2}\rho A C_D|U-\dot X|(U-\dot X)$ (\cite{deLangre2008Wind}, Eq.~5) drops the damping a branch's own motion induces, a simplification the graphics literature makes deliberately on the grounds that it is absorbed into $\zeta$~\cite{Diener2009Wind}. In Sitka spruce, that discarded share is roughly 40\% of total damping, and a further 50\% comes from crowns clashing with neighbours~\cite{Milne1991Dynamics}, a mechanism absent from an isolated tree, so $\zeta$ here is a property of the scene rather than of the wood. And because real damping rises with mean wind speed through streamlining~\cite{Vogel1989Drag,Moore2004Sway} while ours is fixed across the three conditions, a method could recover $\zeta$ perfectly here and still be wrong on real footage. We regard this as a limitation of the testbed, not of the methods evaluated on it.

\paragraph{Splits and Conditions.} The testbed provides a fixed monocular training camera, held-out azimuths at $\pm5/15/30/60^{\circ}$, a temporal-extrapolation tail, and three wind conditions at fixed geometry (differing in mean speed $v_0$), which is usesd to test generalization to unseen forcing.

\paragraph{Generation Caveats.} Two aspects of building this testbed proved consequential. The per-bone oscillators must be integrated to steady state \emph{before} the recorded window opens: starting from $\theta=\dot\theta=0$ at the first rendered frame leaves a transient with $\tau=1/(\zeta\omega)$ reaching $\approx$5.7\,s, which contaminates roughly the first 135 of 300 training frames and biases any $(\hat\omega,\hat\zeta)$ fit to them. We therefore integrate and discard a $1.5\tau_{\max}$ lead-in. Separately, the generator's parameters govern mesh validity far less intuitively than their names suggest: an armature shallower than the geometry hierarchy leaves the deepest level generated but never skinned, collapsing the animated fraction to 1--2\% of a nominally larger bone count.

\subsection{Physically Parameterized Deformation Prior}
\label{sec:method:prior}

Both priors act on one canonical Gaussian set~\cite{Kerbl20233DGS}, optimized photometrically from the monocular training video with gsplat~\cite{Ye2025gsplat} as the rasterizer, each Gaussian holding position, rotation, scale, opacity, and diffuse color (we omit spherical harmonics). The initial positions are read directly off the generator's tree mesh, in the pose it holds at the first training frame, one Gaussian per vertex. Gaussians are hard-assigned to $K$ rigid parts via $k$-means on canonical position (\cref{fig:pipeline}). Each part $p$ carries a learnable $(\omega_p,\zeta_p,$ force gain, swing axis, pivot$)$, driven by the scene's \emph{actual observed} wind $v(t)$, available at test time, like an anemometer reading, never by the ground truth of \cref{sec:method:testbed}, reserved strictly for evaluation. Each frame, $\theta_p(t)$ comes from differentiable RK4 integration of \cref{eq:oscillator}, and a rigid rotation by $\theta_p(t)$ about the part's pivot/axis is applied to \emph{both} position and orientation of its Gaussians -- position-only would misalign elongated, branch-aligned Gaussians as they swing.

\begin{figure*}[!p]
  \centering
  \includegraphics[height=0.86\textheight]{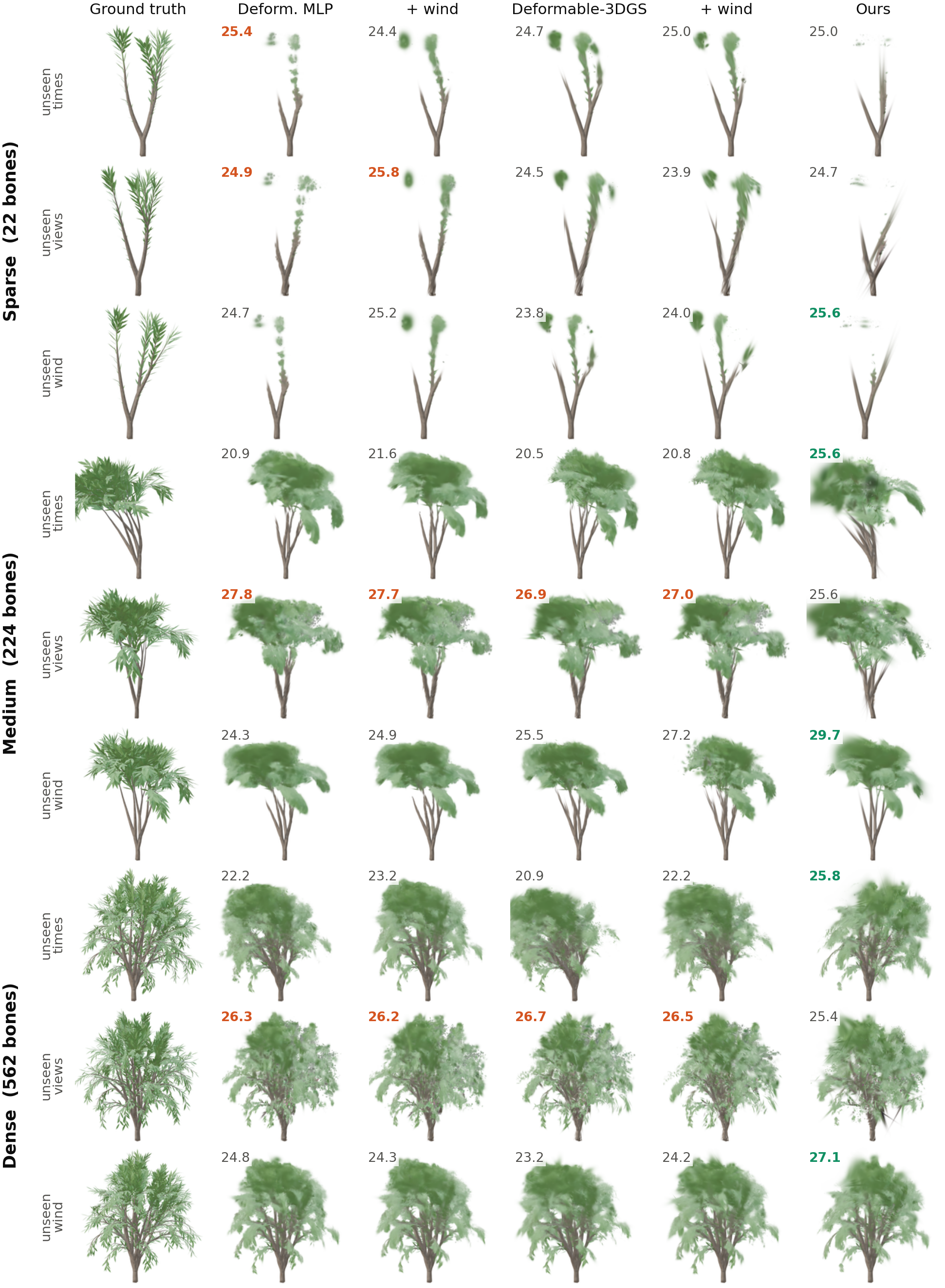}
  \caption{\textbf{Qualitative Results across Three Trees and Three Axes of Generalization.} Every model is trained on one wind condition, one camera and the first 300 frames. \emph{Unseen times} are frames beyond that window, \emph{unseen views} are held-out azimuths, and \emph{unseen wind} is a condition never trained on ($0.71v_0$), evaluated beyond the training window, since at training timestamps a time-indexed field need only replay. Numbers are per-panel PSNR (dB), \textcolor{good}{green} where ours is best and \textcolor{bad}{orange} where a baseline beats it; each panel shows the frame whose ours-minus-baseline gap is closest to that split's mean.}
  \label{fig:resultspage}
\end{figure*}

\section{Experimental Results}
\label{sec:results}

\begin{figure}[t]
  \centering
  \includegraphics[width=\linewidth]{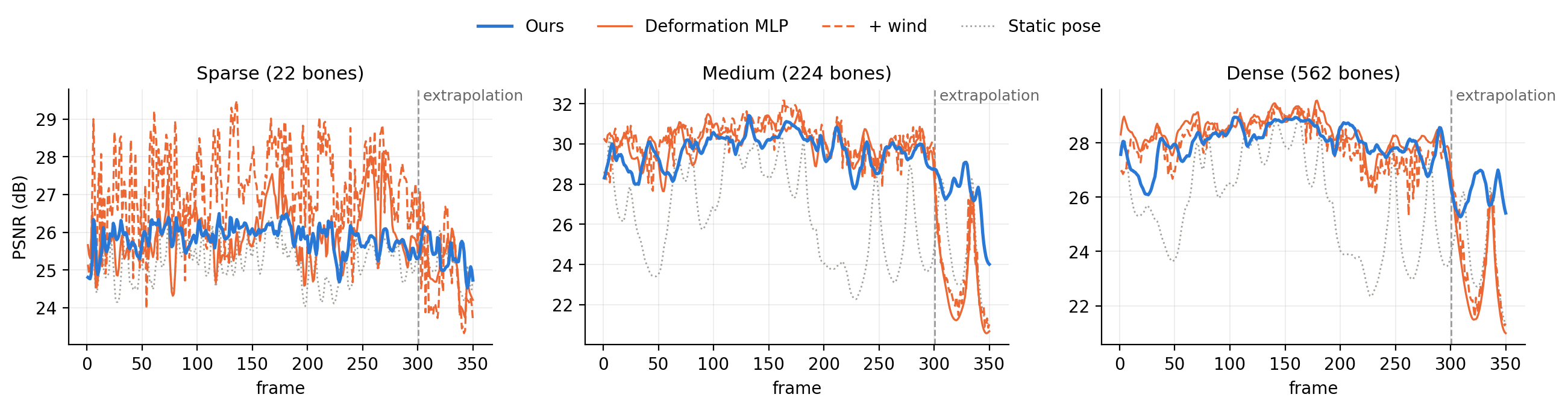}
  \caption{\textbf{The collapse is located exactly at the boundary.} Per-frame PSNR at the training camera and the training wind, so the only thing that changes at frame 301 is that the timestamps leave the fitted window. On the medium and dense trees, the Deformation MLP tracks ours inside the window, wind-conditioned or not, and then falls away at once past it, below even the undeformed canonical pose, while ours continues. The sparse tree separates nothing here either.}
  \label{fig:temporal}
\end{figure}

\paragraph{Setup.} We train from a single fixed camera on 300 frames of one wind condition, and evaluate on held-out azimuths, a 50-frame extrapolation tail, and the three wind conditions of \cref{sec:method:testbed}. Both priors share the same canonical Gaussian initialization and optimizer. Every result is checked across three independent training configurations: a single wind condition; jointly with $K{=}32$ parts; jointly with $K{=}96$. Results are reported on all three tree instances, ordered by skeleton complexity: sparse (22 animated bones), medium (224) and dense (562).

\paragraph{Baselines.}
\label{sec:results:baselines}
\emph{Deformation MLP} is a Fourier-encoded MLP over canonical position and time (6 and 4 bands, 4 layers of width 128) predicting a per-Gaussian position offset: the Deformable-3DGS design~\cite{Yang2024Deformable} reduced to its position head. We also run the full field, which is deeper and wider (10 and 6 bands, 8 layers of width 256 with a skip connection) and predicts per-Gaussian rotation and scale offsets as well. Neither is conditioned on wind as published, so to compare against them fairly we adapt both to receive exactly the signal our prior receives. Three details make the adaptation an opponent rather than a foil. The wind enters as a \emph{history} window rather than an instantaneous $v(t)$: a branch's response depends on its past through the time constant $\tau = 1/(\zeta\omega)$, which spans 1.3--5.7\,s over our band, so we feed the preceding 6\,s subsampled to 48 values. The wind is standardized once by the \emph{training} condition's statistics, frozen into the checkpoint, since standardizing each condition by its own would erase the amplitude difference the transfer test measures. And we feed raw wind speed rather than $v^2-v_0^2$, which would hand the network the drag law we are asking it to discover. Within each architecture, the wind-conditioned and unconditioned variants share encodings, width, depth, zero-initialized output heads, optimizer and training condition, so only the wind input differs. The undeformed canonical pose is reported throughout as a floor.

\subsection{Qualitative Results}
\label{sec:results:qual}

\cref{fig:resultspage} shows the comparison qualitatively, and the pattern separates by axis rather than by tree. On \emph{unseen views}, where the test is appearance alone, every learned baseline beats ours on all three trees: the rigid-part constraint buys nothing when only the camera moves, and costs fidelity. On \emph{unseen times} and \emph{unseen wind}, the ordering inverts on the medium and dense trees, and what fails is not fine detail. The learned baselines collapse into a diffuse canopy that has lost the silhouette, including the full-capacity Deformable-3DGS~\cite{Yang2024Deformable} variant given the observed wind, and their PSNR drops below that of the undeformed pose. The sparse tree separates nothing: every method reconstructs it poorly, which is a property of a 2097-Gaussian scene rather than of any prior. The three subsections below quantify each axis in turn.

\subsection{Reconstruction Quality under Extrapolation}
\label{sec:results:psnr}

\cref{tab:psnr} compares priors on the training wind condition, and \cref{fig:temporal} resolves the same comparison frame by frame. On the medium tree, the physical prior trades a modest amount on held-out azimuths for a clear gain under temporal extrapolation, where the Deformation MLP has no mechanism to keep behaving correctly beyond the frames it was fit to. The signs differ in a way worth stating plainly: the Deformation MLP \emph{loses} 4.81\,dB moving from held-out views to the extrapolation tail, whereas the physical prior \emph{gains} 1.14\,dB, which is what one expects if it has recovered dynamics that remain valid outside the fitting window rather than an interpolant of it. The \emph{time} column tells the same story on all three trees: every learned baseline ends below the undeformed pose, and ours stays above it.

The sparse tree does not reproduce this contrast. Its Deformation MLP degrades by only 0.43\,dB under extrapolation, leaving almost no headroom, and the two priors finish within 0.41\,dB of each other. With 22 bones and little occluded canopy motion, that tree is simply too static to separate the two hypotheses. We therefore read the extrapolation gain as concentrated where canopy motion is complex, rather than as a constant benefit of the physical prior.

\begin{table}[t]
  \centering
  \small
  \resizebox{\linewidth}{!}{%
  \begin{tabular}{lcccccc}
    \toprule
    & \multicolumn{2}{c}{Sparse (22)} & \multicolumn{2}{c}{Medium (224)} & \multicolumn{2}{c}{Dense (562)} \\
    \cmidrule(lr){2-3}\cmidrule(lr){4-5}\cmidrule(lr){6-7}
    & view & time & view & time & view & time \\
    \midrule
    Static pose & 24.99 & 25.16 & 25.35 & 24.70 & 25.08 & 24.09 \\
    Deformation MLP & 25.45 & 25.02 & 27.88 & 23.07 & \textbf{26.59} & 23.19 \\
    \;+\,wind & \textbf{26.39} & 24.75 & \textbf{27.93} & 23.53 & 26.44 & 23.67 \\
    Deformable-3DGS & 26.01 & 24.25 & 27.66 & 23.07 & 26.58 & 22.65 \\
    \;+\,wind & 25.49 & 24.53 & 27.75 & 23.56 & 26.49 & 23.31 \\
    Ours & 25.24 & \textbf{25.42} & 26.37 & \textbf{27.50} & 25.66 & \textbf{26.14} \\
    \bottomrule
  \end{tabular}}
  \caption{\textbf{Two axes of generalization, all three trees.} PSNR (dB). \emph{view}: held-out azimuths at the training wind and inside the training window, so only the camera is new. \emph{time}: frames beyond the training window, at the training wind and camera. ``Static pose'' is the undeformed canonical pose, the reference a deformation model must clear to be doing anything at all. Best per column in bold.}
  \label{tab:psnr}
\end{table}

\subsection{Recovered Natural Frequencies}
\label{sec:results:omega}

The dataset carries no per-Gaussian bone label, so we match a learned part to ground truth by \emph{dynamics}: both solve the same ODE (\cref{eq:oscillator}) under the same forcing, so a correctly-recovered part reproduces its bone's $\theta(t)$. We correlate each part's $\theta_p(t)$ against every bone's, take the best match, and require $|\text{corr}|\!\geq\!0.7$ to trust it.

Taken at face value, the numbers look strong on the medium tree: log-log $R^2$ of $0.931$ (single condition, 30/32 parts trusted), $0.944$ ($K{=}32$, 31/32) and $0.963$ ($K{=}96$, 94/96), with median relative error 1.6--2.4\%.

\paragraph{Null Control.} Matching a part to a bone by trajectory correlation and then reading $\omega_{\text{GT}}$ off the matched bone is a nearest-neighbour lookup against whatever frequency grid the tree happens to provide, and it can be passed without learning anything. We therefore ran the protocol on an \emph{untrained} model whose $\omega_p$ are drawn log-uniformly over the ground-truth range, with $\zeta$ at its initialization, integrated by the same solver and pushed through the same matcher (8 random draws, \cref{tab:null}). On the medium tree, this null scores median relative error 1.77--1.81\% at $R^2 = 0.996$, matching every part: within 0.2 points of the trained model on error, and better on $R^2$ and on parts matched. The reason is density. That tree's 224 bones tile 0.35--1.5\,Hz so finely that an arbitrary frequency lies within 0.40\% (median) of some bone, well inside the 1.6--2.4\% the trained model reports. On this instance, the metric carries no information about whether recovery occurred. The dense tree makes the point sharper still. Its 562 bones tile the axis to within 0.15\%, the null reaches 1.47--1.56\% at $R^2 = 0.997$--$0.998$, and the trained model scores 1.8--2.4\%: on that tree the trained model is \emph{worse} than random frequencies pushed through the same matcher. Degeneracy of this metric scales with skeleton density, and it is total by 562 bones.

The sparse tree is the opposite case, and it is the one that survives. Its 22 bones leave a median nearest-bone gap of 5.40\%, so the null only reaches 4.51--4.56\%, and the jointly-trained model's 1.8\% beats it by $2.5\times$. Its single-condition run (4.3\%) does not beat the null and we do not claim it. We therefore retract frequency recovery as a headline result on dense trees, and restate it in the only form the evidence supports: on the one tree whose bones are sparse enough for the metric to discriminate, joint multi-condition training recovers natural frequencies substantially better than chance, and on the two dense trees our protocol cannot tell whether recovery occurred at all. This is a weaker claim than the numbers first appear to license, and it is why we treat a random-parameter null as a required control rather than an afterthought.

\begin{table}[t]
  \centering
  \small
  \resizebox{\linewidth}{!}{%
  \begin{tabular}{llcc}
    \toprule
    Tree & Model & Median rel.\ err. & log-log $R^2$ \\
    \midrule
    \multirow{2}{*}{Sparse (22 bones)} & Untrained null & 4.56\% & 0.981 \\
     & Trained ($K{=}32$) & \textbf{1.80\%} & 0.810 \\
    \midrule
    \multirow{2}{*}{Medium (224 bones)} & Untrained null & 1.77\% & 0.996 \\
     & Trained ($K{=}32$) & \textbf{1.60\%} & 0.944 \\
    \midrule
    \multirow{2}{*}{Dense (562 bones)} & Untrained null & \textbf{1.56\%} & 0.998 \\
     & Trained ($K{=}32$) & 2.40\% & 0.843 \\
    \bottomrule
  \end{tabular}}
  \caption{\textbf{Frequency recovery against an untrained null.} The null draws each $\omega_p$ log-uniformly over the ground-truth range and pushes it through the same solver and the same trajectory matcher as the trained model; entries are the median of 8 draws. Median relative error is taken over matched parts (lower is better); $R^2$ is that of $\log\hat\omega_p$ against $\log\omega_p$. Bold marks the better of each pair on median relative error.}
  \label{tab:null}
\end{table}

Damping is not recovered. Against a ground truth that is a \emph{constant} 0.08, $\hat\zeta_p$ has median 0.125--0.142 on the two canopy trees and 0.095--0.102 on the sparse tree, and the error is not a uniform offset but a systematic function of frequency: the model fits a nearly frequency-independent decay \emph{rate} $\hat\gamma_p=\hat\zeta_p\hat\omega_p$ in place of a damping \emph{ratio}, reading high on slow parts and below truth on fast ones. A benchmark intending to test damping recovery should vary $\zeta$ across bones and excite a transient on purpose; ours does neither.

\begin{table*}[t]
  \centering
  \small
  \begin{tabular}{llcccccc}
    \toprule
    & & & \multicolumn{2}{c}{Deformation MLP} & \multicolumn{2}{c}{Deformable-3DGS} & \\
    \cmidrule(lr){4-5}\cmidrule(lr){6-7}
    Split & Tree / unseen wind & Static & no wind & +\,wind & no wind & +\,wind & Ours \\
    \midrule
    \multirow{6}{*}{\shortstack[l]{Training\\timestamps}}
      & Sparse (22 bones), $0.84v_0$  & 25.31 & 25.90 & 27.04 & \textbf{27.10} & 27.07 & 25.84 \\
      & Sparse (22 bones), $0.71v_0$  & 25.48 & 25.82 & \textbf{26.49} & 26.06 & 26.13 & 25.87 \\
      & Medium (224 bones), $0.84v_0$ & 27.48 & 28.95 & 29.65 & 29.48 & 29.70 & \textbf{30.12} \\
      & Medium (224 bones), $0.71v_0$ & 28.29 & 28.07 & 28.76 & 28.34 & 28.69 & \textbf{30.43} \\
      & Dense (562 bones), $0.84v_0$  & 26.53 & 28.27 & \textbf{28.32} & 27.90 & 27.79 & 28.22 \\
      & Dense (562 bones), $0.71v_0$  & 27.14 & 27.53 & 27.71 & 27.13 & 27.19 & \textbf{28.46} \\
\midrule
    \multirow{6}{*}{\shortstack[l]{Beyond the\\training window}}
      & Sparse (22 bones), $0.84v_0$  & 25.26 & 25.03 & 25.00 & 24.26 & 24.56 & \textbf{25.56} \\
      & Sparse (22 bones), $0.71v_0$  & 25.40 & 25.13 & 25.31 & 24.33 & 24.66 & \textbf{25.68} \\
      & Medium (224 bones), $0.84v_0$ & 25.64 & 23.50 & 24.00 & 23.39 & 24.08 & \textbf{28.58} \\
      & Medium (224 bones), $0.71v_0$ & 26.54 & 23.95 & 24.47 & 23.75 & 24.59 & \textbf{29.26} \\
      & Dense (562 bones), $0.84v_0$  & 24.80 & 23.51 & 24.02 & 22.83 & 23.59 & \textbf{26.66} \\
      & Dense (562 bones), $0.71v_0$  & 25.55 & 23.85 & 24.34 & 23.06 & 23.90 & \textbf{27.14} \\
\bottomrule
  \end{tabular}
  \caption{\textbf{Zero-shot transfer to unseen wind.} PSNR (dB); every model is fit to $v_0$ only and evaluated on two conditions it never saw. The two blocks differ solely in which frames are scored: training timestamps (top) and frames beyond the training window (bottom). ``Static'' is the undeformed canonical pose, the reference a deformation model must clear to be doing anything at all. Deformable-3DGS columns are the full-capacity field of \cite{Yang2024Deformable}. Best per row in bold.}
  \label{tab:transfer}
\end{table*}

\subsection{Zero-Shot Transfer to Unseen Wind Conditions}
\label{sec:results:transfer}

Whether recovered parameters are physical rather than merely fitted is best tested by extrapolating them to wind never seen in training. We train on one condition and evaluate zero-shot on the other two (\cref{tab:transfer}). An earlier version of this test compared response-amplitude \emph{ratios} across conditions against the drag law's $(v_2/v_1)^2$; we found it vacuous, since the conditions' wind signals are an exact rescaling of one turbulence realization, which, combined with the linearity of \cref{eq:oscillator}, makes the ratio law hold for \emph{any} parameters, including an untrained model's.

\paragraph{Wind Conditioning and Capacity.} The baselines receive the same wind signal we do (\cref{sec:results:baselines}). Wind conditioning is worth 0.5--0.7\,dB to the Deformation MLP on the medium tree and less reliably elsewhere; extra capacity buys nothing systematic, with Deformable-3DGS landing from $-0.80$ to $+1.19$\,dB relative to the matched-capacity MLP and behind it in 16 of 24 matched comparisons. Neither addition closes the gap under the evaluation that matters.

\paragraph{Evaluation Split.} Frames 1--300 of an unseen condition carry \emph{new wind at old timestamps}. A field indexed by time is still in-distribution in time there and can score well by replaying its fitted trajectory, so this split systematically flatters replay: on it, the wind-conditioned baseline beats us on the dense tree at $0.84v_0$ and on the sparse tree at both conditions. Moving to frames beyond the training window changes every cell. Every learned baseline falls \emph{below the static, undeformed pose} in all 24 cells, by 0.10 to 2.79\,dB, while ours stays above it in all six and beats the strongest baseline in every cell by 0.38 to 4.68\,dB. A model worse than not moving at all has transferred nothing; it is outside the window whose motion it memorized.

\cref{tab:margin} reports the one control \cref{tab:transfer} does not, the training condition itself: our prior is behind the wind-conditioned baseline there on all three trees, and the margin then improves as the wind moves away from training, by $+0.78$ to $+2.27$\,dB overall, monotonically except for a $0.08$\,dB dip on the dense tree at $0.84v_0$. The sparse tree is an exception to the magnitude, not the trend: it starts 1.42\,dB behind and gains only 0.81. On the medium tree, we are ahead of both baselines on both held-out conditions even at training timestamps, by 1.17 and 2.36\,dB over the unconditioned MLP and by 0.46 and 1.67\,dB over its wind-conditioned variant, and the widening with divergence is the qualitative signature the argument predicts. We report this candidly: weaker wind means less motion to explain, so the static floor is already close. The sparse tree cannot support this test at all. Beyond the training window, the MLP variants and ours all sit within 0.3\,dB of the undeformed pose, with the higher-capacity field 0.7 to 1.1\,dB below it, and at training timestamps the unconditioned MLP nominally edges us at $0.84v_0$ by under 0.1\,dB, which we read as evidence that the instance lacks the motion needed to discriminate between methods.

\begin{table}[t]
  \centering
  \small
  \resizebox{\linewidth}{!}{%
  \begin{tabular}{lccc}
    \toprule
    Ours $-$ (Deformation MLP $+$\,wind) & $v_0$ & $0.84v_0$ & $0.71v_0$ \\
    \midrule
    Sparse (22 bones)  & $-1.42$ & $-1.21$ & $-0.61$ \\
    Medium (224 bones) & $-0.60$ & $+0.46$ & $+1.67$ \\
    Dense (562 bones)  & $-0.03$ & $-0.11$ & $+0.75$ \\
    \bottomrule
  \end{tabular}}
  \caption{\textbf{Margin against the wind-conditioned baseline.} Each entry is ours minus Deformation MLP\,$+$\,wind in PSNR (dB) at training timestamps, on the training condition $v_0$ and on the two conditions never seen. Positive means ours is ahead.}
  \label{tab:margin}
\end{table}

\subsection{Discussion and Limitations}
\label{sec:results:limitations}

Recovery is limited by the evaluation as much as by the model: frequency recovery clears a random-parameter null on the sparse tree alone, the canopy trees being too finely tiled to discriminate. Our segmentation is spatial $k$-means (\cref{fig:pipeline}), not topology-aware, which plausibly feeds the damping bias. More fundamentally, each part is an independent oscillator rotating rigidly about a fixed pivot, representing neither bending within a part nor coupling between parent and child branches. It also needs the forcing observed, a synchronized $v(t)$ at test time, so real footage needs an anemometer beside the camera. The prior's cost in appearance is $0.2$--$1.6$\,dB on held-out azimuths (\cref{tab:psnr}); no margin here is backed by seed variance; and the testbed has one generator, no real species and no real footage. We also compare only against learned deformation fields; fitting the same oscillators to tracked 2D trajectories would separate what the physics buys from what the renderer costs.

\section{Conclusion}

We test whether a physically parameterized deformation prior recovers a tree's dynamics or merely fits its appearance. On extrapolation, it helps: once evaluation leaves the fitted window, in time or wind speed, every learned baseline falls below an undeformed pose and ours does not. On recovery, the answer is weaker: matching parts to bones is degenerate on dense skeletons, and damping is not recovered.

{
    \small
    \bibliographystyle{ieeenat_fullname}
    \bibliography{main}
}


\end{document}